%% file: main_camera_ready.tex
\documentclass[10pt,twocolumn,letterpaper]{article}

\usepackage{cvpr}
\usepackage{times}
\usepackage{latexsym}
\usepackage{graphicx}
\usepackage{wrapfig}
\usepackage{amsmath,bm}
\usepackage{amssymb}
\usepackage{algorithm}
\usepackage{algpseudocode}
\usepackage{multirow}
\usepackage{pifont}
\usepackage{amssymb}
\usepackage{color}
\usepackage{enumitem}
\usepackage{setspace}
\usepackage{indentfirst}
\usepackage{url}

\usepackage[pagebackref=true,breaklinks=true,letterpaper=true,colorlinks,bookmarks=false]{hyperref}

\newcommand{\x}{{\bf x}}
\newcommand{\bb}{{\bf b}}

\newcommand{\z}{{\bf z}}
\newcommand{\X}{\text{X}}

\newcommand{\btheta}{\bm{\theta}}
\newcommand{\bphi}{\bm{\phi}}
\newcommand{\EE}{{\mathbb{E}}}

\input{math_commands.tex}

\usepackage{enumitem}
\usepackage{setspace}
\setlist[itemize]{leftmargin=5.mm}

\cvprfinalcopy 

\def\cvprPaperID{5596} 

\ifcvprfinal\fi

\begin{document}
\author{Zheng Ding\thanks{Authors contributed equally.} $^{,1,2}$, Yifan Xu$^{*,2}$, Weijian Xu$^{2}$, Gaurav Parmar$^{2}$, Yang Yang$^{3}$, Max Welling$^{3, 4}$, Zhuowen Tu$^{2}$\\
$^{1}$Tsinghua University \quad \quad $^{2}$UC San Diego \quad \quad $^{3}$Qualcomm, Inc. \quad \quad $^{4}$University of Amsterdam\\
}

\title{Guided Variational Autoencoder for Disentanglement Learning}

\maketitle
\thispagestyle{empty}

\begin{abstract}
We propose an algorithm, guided variational autoencoder (Guided-VAE), that is able to learn a controllable generative model by performing latent representation disentanglement learning. The learning objective is achieved by providing signals to the latent encoding/embedding in VAE without changing its main backbone architecture, hence retaining the desirable properties of the VAE. We design an unsupervised strategy and a supervised strategy in Guided-VAE and observe enhanced modeling and controlling capability over the vanilla VAE. In the unsupervised strategy, we guide the VAE learning by introducing a lightweight decoder that learns latent geometric transformation and principal components; in the supervised strategy, we use an adversarial excitation and inhibition mechanism to encourage the disentanglement of the latent variables. Guided-VAE enjoys its transparency and simplicity for the general representation learning task, as well as disentanglement learning. On a number of experiments for representation learning, improved synthesis/sampling, better disentanglement for classification, and reduced classification errors in meta learning have been observed.

\end{abstract}

\section{Introduction}
\label{sec:intro}

The resurgence of autoencoders (AE) \cite{yann1987modeles,bourlard1988auto,hinton1994autoencoders} is an important component in the rapid development of modern deep learning \cite{goodfellow2016deep}. Autoencoders have been widely adopted for modeling signals and images \cite{poultney2007efficient, vincent2010stacked}.
Its statistical counterpart, the variational autoencoder (VAE) \cite{kingma2013auto}, has led to a recent wave of development in generative modeling due to its two-in-one capability, both representation and statistical learning in a single framework. Another exploding direction in generative modeling includes generative adversarial networks (GAN) \cite{goodfellow2014generative}, but GANs focus on the generation process and are not aimed at representation learning (without an encoder at least in its vanilla version). 

Compared with classical dimensionality reduction methods like principal component analysis (PCA) \cite{candes1933robust,Jolliffe2011principal} and Laplacian eigenmaps \cite{belkin2003laplacian}, VAEs have demonstrated their unprecedented power in modeling high dimensional data of real-world complexity. However, there is still a large room to improve for VAEs to achieve a high quality reconstruction/synthesis. Additionally, it is desirable to make the VAE representation learning more transparent, interpretable, and controllable.

In this paper, we attempt to learn a transparent representation by introducing guidance to the latent variables in a VAE. We design two strategies for our Guided-VAE, an unsupervised version (Fig.~\ref{fig:model}.a) and a supervised version (Fig.~\ref{fig:model}.b). The main motivation behind Guided-VAE is to encourage the latent representation to be semantically interpretable, while maintaining the integrity of the basic VAE architecture. Guided-VAE is learned in a multi-task learning fashion. The objective is achieved by taking advantage of the modeling flexibility and the large solution space of the VAE under a lightweight target. Thus the two tasks, learning a good VAE and making the latent variables controllable, become companions rather than conflicts.

In {\bf unsupervised Guided-VAE}, in addition to the standard VAE backbone, we also explicitly force the latent variables to go through a lightweight encoder that learns a deformable PCA. As seen in Fig.~\ref{fig:model}.a, two decoders exist, both trying to reconstruct the input data $\x$:
The main decoder, denoted as $\text{Dec}_{main}$, functions regularly as in the standard VAE \cite{kingma2013auto}; the secondary decoder, denoted as $\text{Dec}_{sub}$, explicitly learns a geometric deformation together with a linear subspace.
In {\bf supervised Guided-VAE}, we introduce a subtask for the VAE by forcing one latent variable to be discriminative (minimizing the classification error) while making the rest of the latent variable to be adversarially discriminative (maximizing the minimal classification error). This subtask is achieved using an adversarial excitation and inhibition formulation. Similar to the unsupervised Guided-VAE, the training process is carried out in an end-to-end multi-task learning manner. The result is a regular generative model that keeps the original VAE properties intact, while having the specified latent variable semantically meaningful and capable of controlling/synthesizing a specific attribute.
We apply Guided-VAE to the data modeling and few-shot learning problems and show favorable results on the MNIST, CelebA, CIFAR10 and Omniglot datasets.

The contributions of our work can be summarized as follows:
{
\begin{itemize}
 \setlength\itemsep{0mm}
 \setlength{\itemindent}{0mm}
\item We propose a new generative model disentanglement learning method by introducing latent variable guidance to variational autoencoders (VAE). Both unsupervised and supervised versions of Guided-VAE have been developed. 
\item In unsupervised Guided-VAE, we introduce deformable PCA as a subtask to guide the general VAE learning process, making the latent variables interpretable and controllable.
\item In supervised Guided-VAE, we use an adversarial excitation and inhibition mechanism to encourage the disentanglement, informativeness, and controllability of the latent variables.
\end{itemize}
}

Guided-VAE can be trained in an end-to-end fashion. It is able to keep the attractive properties of the VAE while significantly improving the controllability of the vanilla VAE.  It is applicable to a range of problems for generative modeling and representation learning.

\section{Related Work}
Related work can be discussed along several directions.

Generative model families such as generative adversarial networks (GAN) \cite{goodfellow2014generative,WGAN} and variational autoencoder (VAE) \cite{kingma2013auto} have received a tremendous amount of attention lately. Although GAN produces higher quality synthesis than VAE, GAN is missing the encoder part and hence is not directly suited for representation learning.
Here, we focus on disentanglement learning by making VAE more controllable and transparent.

Disentanglement learning \cite{mathieu2016disentangling,szabo2017challenges,hu2018disentangling,achille2018emergence,gonzalez2018image,jha2018disentangling} recently becomes a popular topic in representation learning. Adversarial training has been adopted in approaches such as \cite{mathieu2016disentangling,szabo2017challenges}.
Various methods \cite{peng2017reconstruction,kim2018disentangling,lin2019exploring} have imposed constraints/regularizations/supervisions to the latent variables, but these existing approaches often involve an architectural change to the VAE backbone and the additional components in these approaches are not provided as secondary decoder for guiding the main encoder.
A closely related work is the $\beta$-VAE \cite{higgins2017beta} approach in which a balancing term $\beta$ is introduced to control the capacity and the independence prior. $\beta$-TCVAE \cite{chen2018isolating} further extends $\beta$-VAE by introducing a total correlation term.

\begin{figure*}[!htp]
\begin{center}
\scalebox{0.9}{
\begin{tabular}{cc}
\hspace{-2mm}
\includegraphics[width=0.5\textwidth]{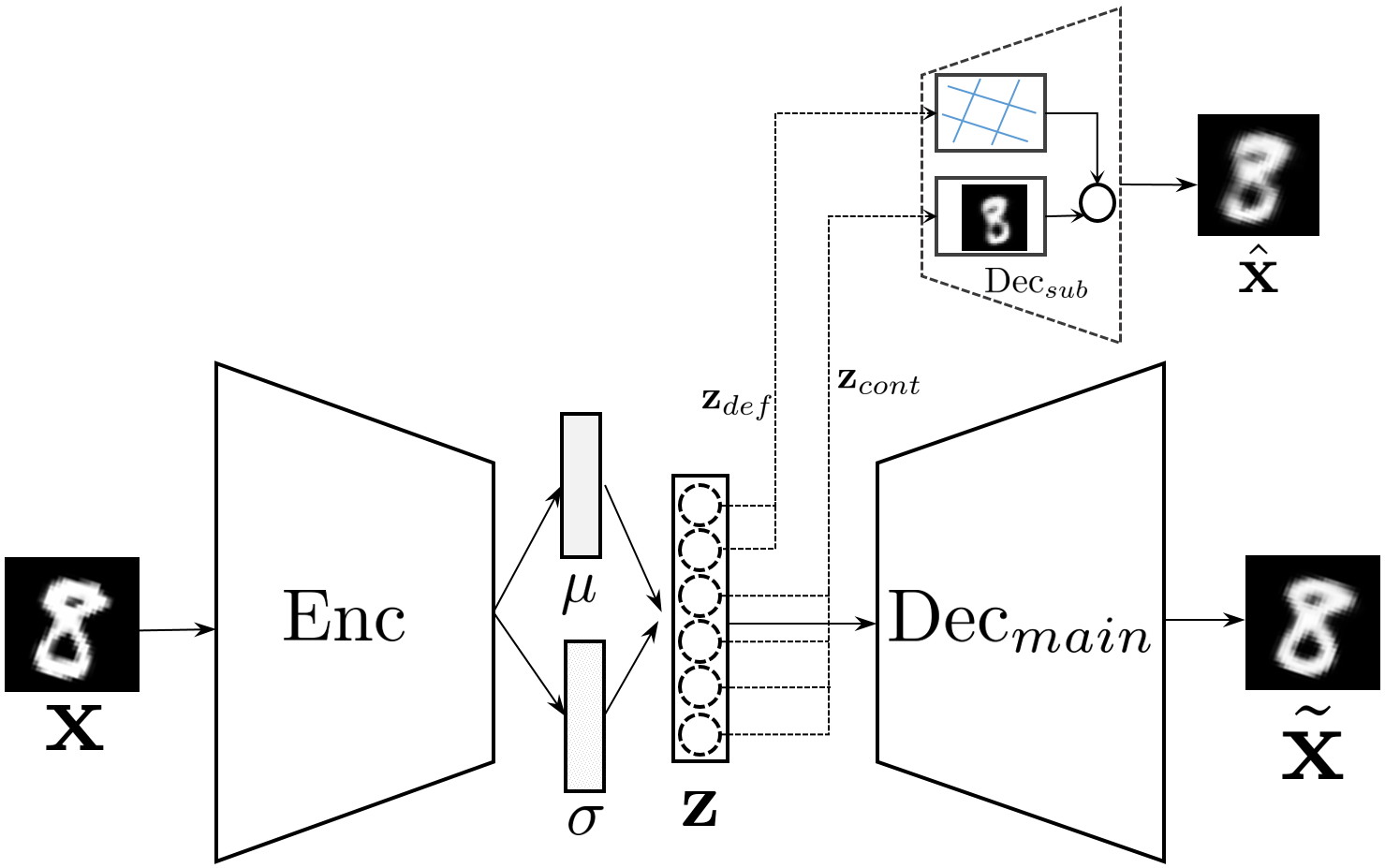}  &
\hspace{5mm}
\includegraphics[width=0.5\textwidth]{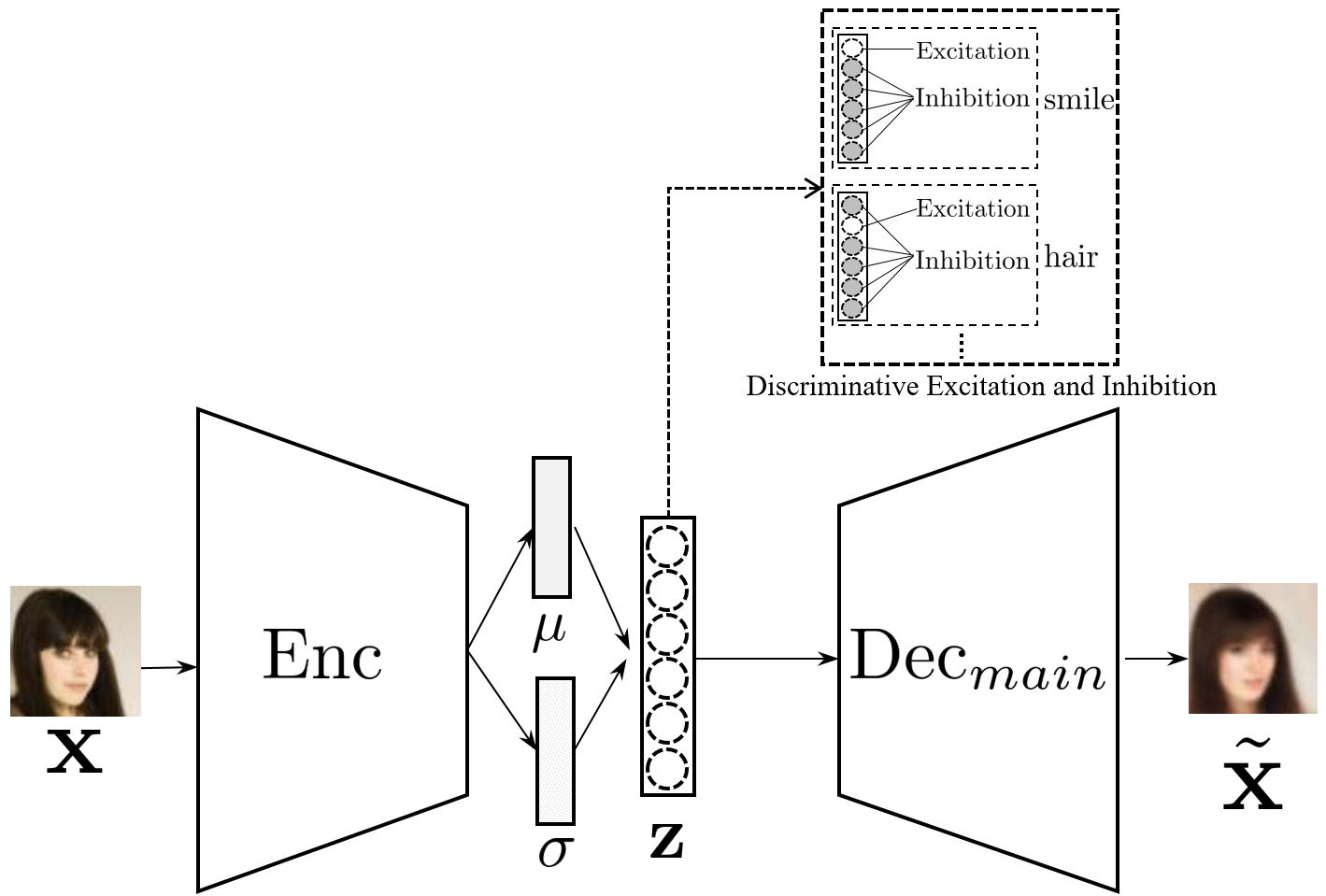}\\
(a) Unsupervised Guided-VAE  &
(b) Supervised Guided-VAE\\
\hspace{-2mm}
\end{tabular}
}
\caption{Model architecture for the proposed Guided-VAE algorithms.}
\label{fig:model}
\vspace{-3mm}
\end{center}
\end{figure*}

From a different angle, principal component analysis (PCA) family \cite{candes1933robust,Jolliffe2011principal,candes2011robust} can also be viewed as representation learning. Connections between robust PCA \cite{candes2011robust} and VAE \cite{kingma2013auto} have been observed \cite{dai2018connections}. Although being a widely adopted method, PCA nevertheless has limited modeling capability due to its linear subspace assumption. To alleviate the strong requirement for the input data being pre-aligned, RASL \cite{peng2012rasl} deals with unaligned data by estimating a hidden transformation to each input.
Here, we take advantage of the transparency of PCA and the modeling power of VAE by developing a sub-encoder (see Fig. \ref{fig:model}.a), deformable PCA, that guides the VAE training process in an integrated end-to-end manner. After training, the sub-encoder can be removed by keeping the main VAE backbone only.

To achieve disentanglement learning in supervised Guided-VAE, we encourage one latent variable to directly correspond to an attribute while making the rest of the variables uncorrelated. This is analogous to the excitation-inhibition mechanism \cite{murphy2003multiplicative, yizhar2011neocortical}
or the explaining-away \cite{wellman1993explaining} phenomena. Existing approaches \cite{liu2018detach,lin2019exploring} impose supervision as a conditional model for an image translation task, whereas our supervised Guided-VAE model targets the generic generative modeling task by using an adversarial excitation and inhibition formulation. This is achieved by minimizing the discriminative loss for the desired latent variable while maximizing the minimal classification error for the rest of the variables. 
Our formulation has a connection to the domain-adversarial neural networks (DANN) \cite{ganin2016domain}, but the two methods differ in purpose and classification formulation. Supervised Guided-VAE is also related to the adversarial autoencoder approach \cite{makhzani2016adversarial}, but the two methods differ in the objective, formulation, network structure, and task domain. In \cite{ilse2019diva}, the domain invariant variational autoencoders method (DIVA) differs from ours by enforcing disjoint sectors to explain certain attributes.

Our model also has connections to the deeply-supervised nets (DSN) \cite{lee2015deeply}, where intermediate supervision is added to a standard CNN classifier. There are also approaches \cite{engel2018latent,bojanowski2018optimizing} in which latent variables constraints are added, but they have different formulations and objectives than Guided-VAE. Recent efforts in fairness disentanglement learning \cite{creager2019flexibly,song2018learning} also bear some similarity, but there is still a large difference in formulation.

\section{Guided-VAE Model}
\label{Guided-VAE}
In this section, we present the main formulations of our Guided-VAE models. The unsupervised Guided-VAE version is presented first, followed by introduction of the supervised version.

\subsection{VAE}

Following the standard definition in variational autoencoder (VAE) \cite{kingma2013auto}, a set of input data is denoted as $\X=(\x_1,...,\x_n)$ where $n$ denotes the number of total input samples. The latent variables are denoted by vector $\z$. The encoder network includes network and variational parameters $\bphi$  that produces variational probability model $q_{\bphi}(\z|\x)$. The decoder network is parameterized by $\btheta$ to reconstruct sample $\tilde{\x}=f_{\btheta}(\z)$. The log likelihood $\log p(\x)$ estimation is achieved by maximizing the Evidence Lower BOund (ELBO) \cite{kingma2013auto}:
\begin{equation}
\begin{aligned}
    ELBO(\btheta, \bphi; \x) &= \EE_{q_{\bphi}(\z|\x)} [\log(p_{\btheta}(\x|\z))] \\
    &- \KL(q_{\bphi}(\z|\x) || p(\z)).
\label{eq:ELBO}
\end{aligned}
\end{equation}

The first term in Eq. (\ref{eq:ELBO}) corresponds to a reconstruction loss $\int q_{\bphi}(\z|\x) \times ||\x-f_{\btheta}(\z)||^2  d\z$ (the first term is the \emph{negative} of reconstruction loss between input $\x$ and reconstruction $\tilde{\x}$) under Gaussian parameterization of the output.
The second term in Eq. (\ref{eq:ELBO}) refers to the KL divergence between the variational distribution $q_{\bphi}(\z|\x)$ and the prior distribution $p(\z)$. 
The training process thus tries to optimize:
\begin{equation}
    \max_{\btheta, \bphi} \left\{\sum_{i=1}^n ELBO(\btheta, \bphi; \x_i)\right\}.
\label{eq:VAE}
\end{equation}
\vspace{-6.5mm}
\subsection{Unsupervised Guided-VAE }
In our unsupervised Guided-VAE, we introduce a deformable PCA as a secondary decoder to guide the VAE training. An illustration can be seen in Fig. \ref{fig:model}.a. This secondary decoder is called $\text{Dec}_{sub}$.
Without loss of generality, we let $\z=(\z_{def}, \z_{cont})$. $\z_{def}$ decides a deformation/transformation field, e.g. an affine transformation denoted as $\tau(\z_{def})$. $\z_{cont}$ determines the content of a sample image for transformation. The PCA model consists of $K$ basis $B=(\bb_1,...,\bb_K)$. We define a deformable PCA loss as:
\begin{equation}
\begin{aligned}
    & \gL_{DPCA}(\bphi, B)\\
    &= \sum_{i=1}^n  \EE_{q_{\bphi}(\z_{def},\z_{cont}|\x_i)}\left[ ||\x_i- \tau(\z_{def}) \circ (\z_{cont} B^T) ||^2 \right] \\
    &+ \sum_{k,j\ne k} (\bb_{k}^T \bb_{j})^2,
\label{eq:DPCA}
\end{aligned}
\end{equation}
where $\circ$ defines a transformation (affine in our experiments) operator decided by $\tau(\z_{def})$ and $\sum_{k,j\ne k} (\bb_{k}^T \bb_{j})^2$ is regarded as the orthogonal loss. A normalization term $\sum_{k} (\bb_{k}^T \bb_{k}-1)^2$ can be optionally added to force the basis to be unit vectors. We follow the spirit of the PCA optimization and a general formulation for learning PCA can be found in \cite{candes2011robust}.

To keep the simplicity of the method we learn a fixed basis $B$ and one can also adopt a probabilistic PCA model \cite{tipping1999probabilistic}. Thus, learning unsupervised Guided-VAE becomes:
\begin{equation}
\begin{aligned}
    \max_{\btheta, \bphi, B} \left \{  \sum_{i=1}^n  ELBO(\btheta, \bphi; \x_i)   - \gL_{DPCA}(\bphi, B) \right \}.
\label{eq:ungvae}
\end{aligned}
\end{equation}
The affine matrix described in our transformation follows implementation in \cite{jaderberg2015spatial}:

\begin{equation}
 A_\theta=
\left[ {
\begin{array}{ccc}
\theta_{11} & \theta_{12} & \theta_{13}\\
\theta_{21} & \theta_{22} & \theta_{23}
\end{array}
}\right]
\label{eqn:attention}
\end{equation}
The affine transformation includes translation, scale, rotation and shear operation. We use different latent variables to calculate different parameters in the affine matrix according to the operations we need.

\begin{figure*}[!htp]
\begin{center}
\begin{tabular}{ccccc}

\includegraphics[width=1\textwidth]{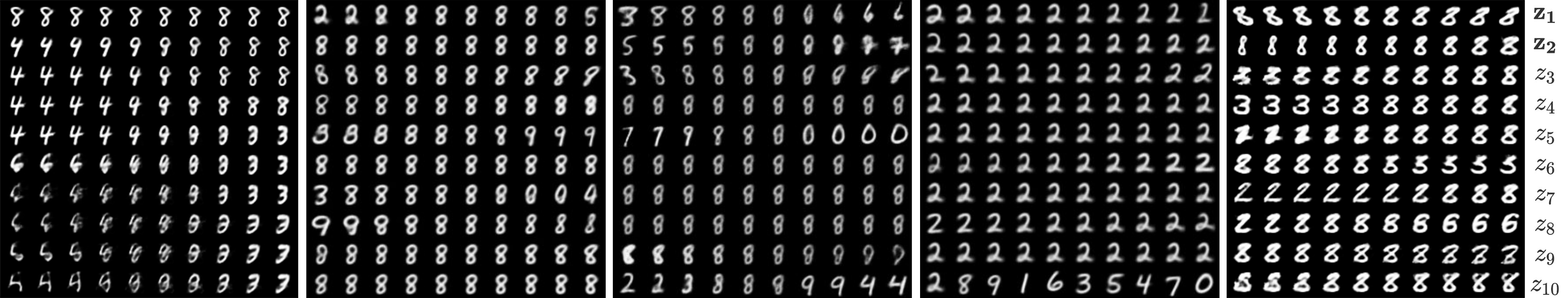}\\
\hspace{1.0em} (a)VAE \hspace{6.0em} (b) $\beta$-VAE \hspace{4.0em} (c) CC$\beta$-VAE \hspace{4.0em} (d) JointVAE \hspace{2.0em} (e) Guided-VAE (Ours) 
\end{tabular}

\caption{\small{\textbf{Latent Variables Traversal on MNIST:} Comparison of traversal results from vanilla VAE  \cite{kingma2013auto}, $\beta$-VAE \cite{higgins2017beta},  $\beta$-VAE with
controlled capacity increase (CC$\beta$-VAE), JointVAE \cite{dupont2018learning} and our Guided-VAE on the MNIST dataset. $z_{1}$ and $z_{2}$ in Guided-VAE are controlled.}}
\label{fig:mnist}
\vspace{-3mm}
\end{center}
\end{figure*}

\vspace{-1mm}
\subsection{Supervised Guided-VAE}
\vspace{-1mm}
For training data $\X=(\x_1,...,\x_n)$, suppose there exists a total of $T$ attributes with ground-truth labels.
Let $\z=(z_t, \z_t^{rst})$ where $z_t$ defines a scalar variable deciding the $t$-th attribute and $\z_t^{rst}$ represents remaining latent variables. Let $y_t(\x_i)$ be the  ground-truth label for the $t$-th attribute of sample $\x_i$; $y_t(\x_i) \in \{-1, +1\}$. For each attribute, we use an adversarial excitation and inhibition method with term: 

\begin{equation}
\begin{aligned}
& \gL_{Excitation}(\bphi, t) \\
&= \max_{w_t}\left\{ \sum_{i=1}^n \EE_{q_{\bphi}(z_t|\x_i)} [\log p_{w_t}(y=y_t(\x_i)|z_t)]\right\} ,
\end{aligned}
\end{equation}
where $w_t$ refers to classifier making a prediction for the $t$-th attribute using the latent variable $z_t$.

This is an excitation process since we want latent variable $z_t$ to directly correspond to the attribute label.

Next is an inhibition term.
\begin{equation}
\begin{aligned}
& \gL_{Inhibition} (\bphi, t) \\
&= \max_{C_t} \left\{\sum_{i=1}^n \EE_{q_{\bphi}(\z_t^{rst}|\x_i)} [\log p_{C_t}(y=y_t(\x_i)|\z_t^{rst})] \right\},
\label{eq:inhibition}
\end{aligned}
\end{equation}
where $C_t(\z_t^{rst})$ refers to classifier making a prediction for the $t$-th attribute using the remaining latent variables $\z_t^{rst}$.
$\log p_{C_t}(y=y_t(\x)|\z_t^{rst})$ is a cross-entropy term for minimizing the classification error in Eq. (\ref{eq:inhibition}).
This is an inhibition process since we want the remaining variables $\z_t^{rst}$ as independent as possible to the attribute label in Eq. (\ref{eq:sugvae}) below.
\begin{equation}
\begin{aligned}
    &\max_{\btheta, \bphi} {\bigg\{}  \sum_{i=1}^n  ELBO(\btheta, \bphi; \x_i)  \\
    &+ \sum_{t=1}^T \left[\gL_{Excitation}(\bphi, t) - \gL_{Inhibition} (\bphi,t) \right] {\bigg\}}.
\label{eq:sugvae}
\end{aligned}
\end{equation}

Notice in Eq. (\ref{eq:sugvae}) the minus sign in front of the term $\gL_{Inhibition} (\bphi, t)$ for maximization which is an adversarial term to make $\z_t^{rst}$ as uninformative to attribute $t$ as possible, by pushing the best possible classifier $C_t$ to be the least discriminative.
The formulation of Eq. (\ref{eq:sugvae}) bears certain similarity to that in domain-adversarial neural networks \cite{ganin2016domain} in which the label classification is minimized with the domain classifier being adversarially maximized. Here, however, we respectively encourage and discourage different parts of the features to make the same type of classification. 

\section{Experiments}
\label{Experiments}

In this section, we first present qualitative and quantitative results demonstrating our proposed unsupervised Guided-VAE (Figure \ref{fig:model}a) capable of disentangling latent embedding more favorably than previous disentangle methods \cite{higgins2017beta, dupont2018learning, kim2018disentangling} on MNIST dataset \cite{lecun2010mnist} and 2D shape dataset \cite{dsprites17}. We also show that our learned latent representation improves classification performance in a representation learning setting. Next, we extend this idea to a supervised guidance approach in an adversarial excitation and inhibition fashion, where a discriminative objective for certain image properties is given (Figure \ref{fig:model}b) on the CelebA dataset \cite{liu2015faceattributes}. Further, we show that our method is architecture agnostic, applicable in a variety of scenarios such as image interpolation task on CIFAR 10 dataset \cite{cifar10} and a few-shot classification task on Omniglot dataset \cite{lake2015human}.

\subsection{Unsupervised Guided-VAE}

\subsubsection{Qualitative Evaluation}

We present qualitative results on the MNIST dataset first by traversing latent variables received affine transformation guiding signal in Figure \ref{fig:mnist}. Here, we applied the Guided-VAE with the bottleneck size of 10 (i.e. the latent variables $\z \in \mathbb{R}^{10}$). The first latent variable $z_{1}$ represents the rotation information, and the second latent variable $z_{2}$ represents the scaling information. The rest of the latent variables $\z_{3:10}$ represent the content information. Thus, we present the latent variables as $\z = (\z_{def}, \z_{cont}) = (\z_{1:2}, \z_{3:10})$.

We compare traversal results of all latent variables on MNIST dataset for vanilla VAE  \cite{kingma2013auto}, $\beta$-VAE \cite{higgins2017beta}, JointVAE \cite{dupont2018learning} and our Guided-VAE  ($\beta$-VAE, JointVAE results are adopted from \cite{dupont2018learning}). While $\beta$-VAE cannot generate meaningful disentangled representations for this dataset, even with controlled capacity increased, JointVAE can disentangle class type from continuous factors. Our Guided-VAE disentangles geometry properties rotation angle at $z_{1}$ and stroke thickness at $z_{2}$ from the rest content information $\z_{3:10}$.  

\begin{figure}
\begin{center}
\scalebox{0.85}{
\begin{tabular}{c}
\hspace{1.2em} $\beta$-VAE \hspace{8.6em} FactorVAE\\
\includegraphics[width=0.52\textwidth]{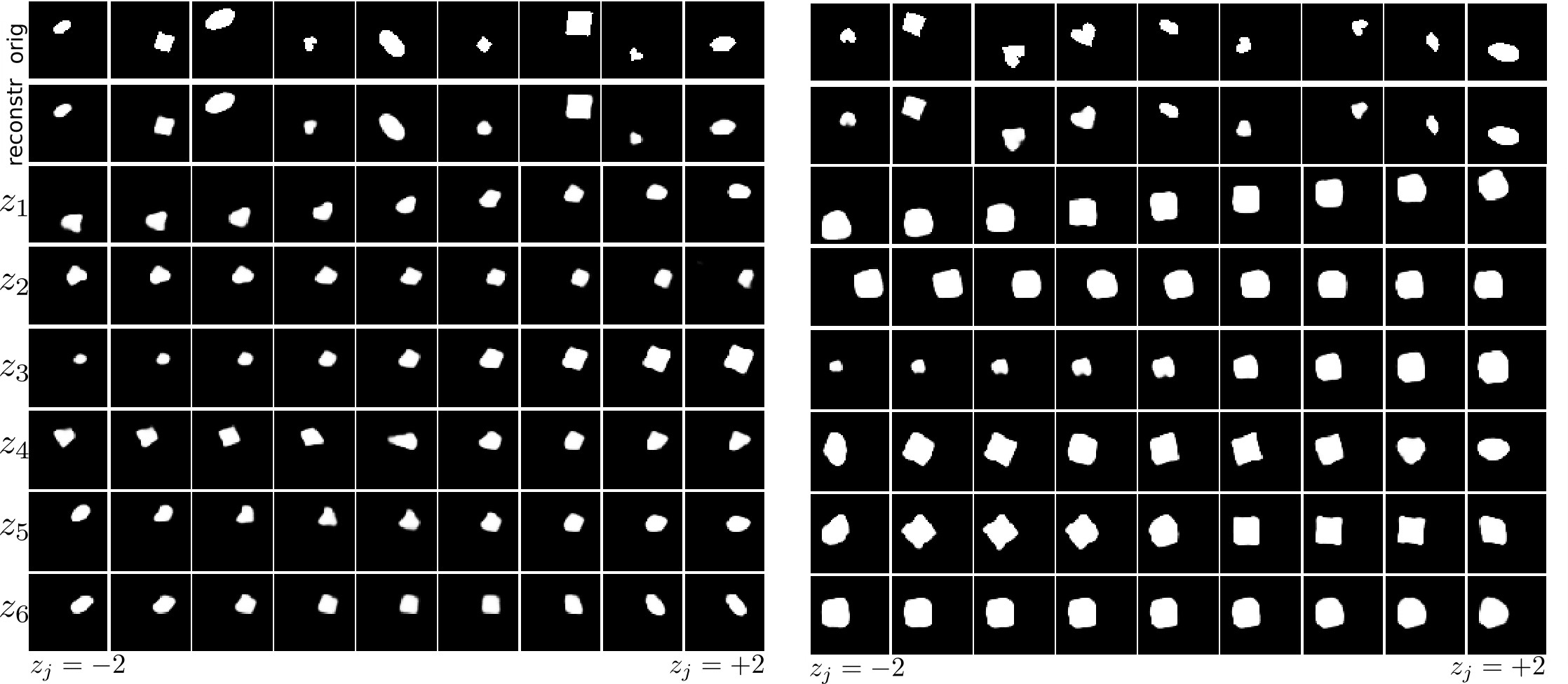}\\
\hspace{4.1em} VAE \hspace{7.2em} Guided-VAE (Ours)\\
\includegraphics[width=0.52\textwidth]{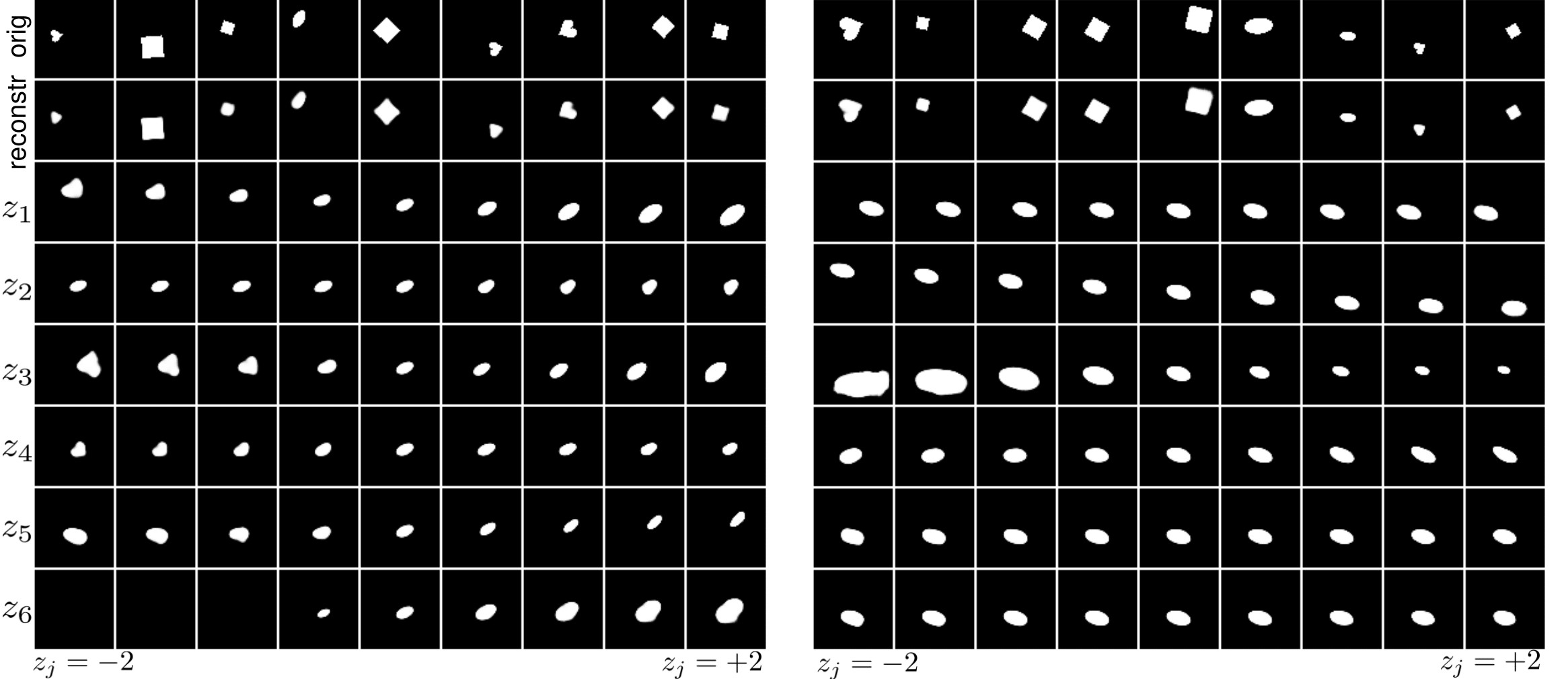}

\end{tabular}
}
\caption{\small \textbf{Comparison of qualitative results on 2D shape.} First row: originals. Second row: reconstructions. Remaining rows: reconstructions of latent traversals across each
latent dimension. In our results, $z_1$ represents the $x$-position information, $z_2$ represents the $y$-position information, $z_3$ represents the scale information and $z_4$ represents the rotation information. }
\label{fig:dsprites}
\vspace{-8mm}
\end{center}
\end{figure}

To assess the disentangling ability of Guided-VAE against various baselines, we create a synthetic 2D shape dataset following \cite{dsprites17, higgins2017beta} as a common way to measure the disentanglement properties of unsupervised disentangling methods. The dataset consists 737,280 images of 2D shapes (heart, oval and square) generated from four ground truth independent latent factors: $x$-position information (32 values), $y$-position information (32 values), scale (6 values) and rotation (40 values). This gives us the ability to compare the disentangling performance of different methods with given ground truth factors. We present the latent space traversal results in Figure \ref{fig:dsprites}, where the results of $\beta$-VAE and FactorVAE are taken from \cite{kim2018disentangling}. Our Guided-VAE learns the four geometry factors with the first four latent variables where the latent variables $\z \in \mathbb{R}^{6} = (\z_{def}, \z_{cont}) = (\z_{1:4}, \z_{5:6})$. We observe that although all models are able to capture basic geometry factors, the traversal results from Guided-VAE are more obvious with fewer factors changing except the target one. 

\begin{figure}[!htp]

\begin{center}
\scalebox{0.9}{
\begin{tabular}{c} 
\includegraphics[width=0.5\textwidth]{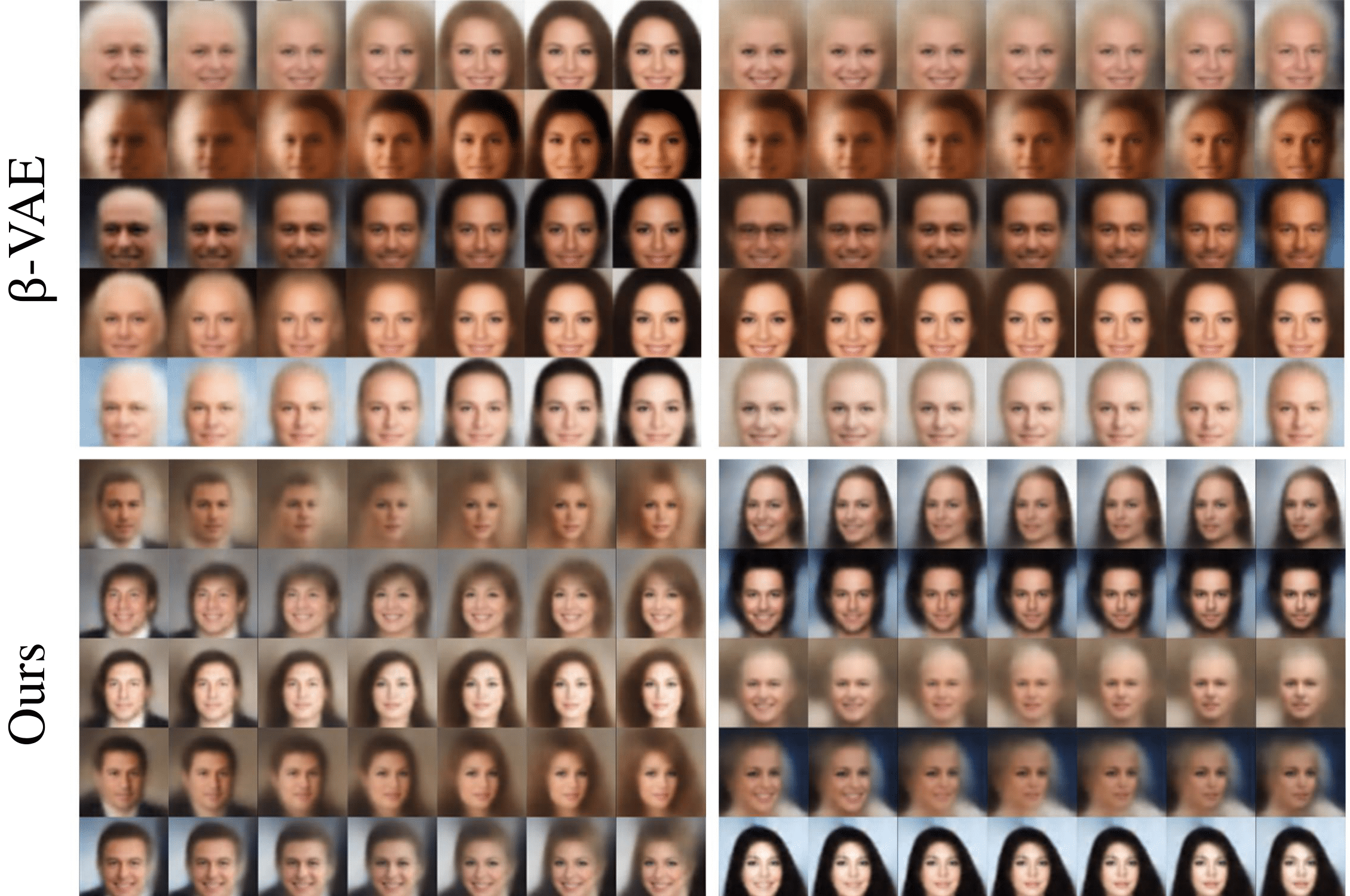}\\
\hspace{1.0em} Gender \hspace{8.0em} Smile
\end{tabular}
}
\caption{\small\textbf{Comparison of Traversal Result learned on CelebA:} Column 1 shows traversed images from male to female. Column 2 shows traversed images from smiling to no-smiling. The first row is from \cite{higgins2017beta} and we follow its figure generation procedure.}
\label{fig:Comparison_CelebA}
\end{center}
\vspace{-5mm}
\end{figure}
\vspace{-3mm}
\subsubsection{Quantitative Evaluation}

We perform two quantitative experiments with strong baselines for disentanglement and representation learning in Table \ref{tab:disentangle_results} and \ref{tab:classification-mnist-methods}. We observe significant improvement over existing methods in terms of {\em disentanglement} measured by Z-Diff score \cite{higgins2017beta}, SAP score \cite{kumar2017variational}, Factor score \cite{kim2018disentangling} in Table \ref{tab:disentangle_results}, and representation {\em transferability} based on classification error in Table \ref{tab:classification-mnist-methods}.

\begin{table}
\begin{center}
\scalebox{0.8}{
\begin{tabular}{l | cccc}
\textbf{Model ($d_\z=6$)} & {Z-Diff $\uparrow$} &  {SAP $\uparrow$} & {Factor $\uparrow$}  \\
\hline
\textbf{\textsc{VAE}   \cite{kingma2013auto}}
            & 78.2 & 0.1696 & 0.4074   \\
\textbf{\textsc{$\beta$-VAE ($\beta$=2)\cite{higgins2017beta}}}
            & 98.1 & 0.1772 & 0.5786  \\
\textbf{\textsc{FactorVAE ($\gamma$=5) \cite{kim2018disentangling}}} 
            & 92.4 & 0.1770 & 0.6134  \\
\textbf{\textsc{FactorVAE ($\gamma$=35) \cite{kim2018disentangling}}} 
            & 98.4 & 0.2717 & 0.7100  \\
\textbf{\textsc{$\beta$-TCVAE ($\alpha$=1,$\beta$=5,$\gamma$=1) \cite{chen2018isolating}}} 
            & 96.8 & 0.4287 & 0.6968  \\
\hline
\textbf{\textsc{Guided-VAE (Ours)}}   
            & \textbf{99.2} & 0.4320 & 0.6660 \\ 
\textbf{\textsc{Guided-$\beta$-TCVAE (Ours)}}   
            & 96.3 & \textbf{0.4477} & \textbf{0.7294} \\ 
\end{tabular}
}
\caption{\small \textbf{Disentanglement:}  Z-Diff score, SAP score, and Factor score over unsupervised disentanglement methods on 2D Shapes dataset. [$\uparrow$ means higher is better]}
\label{tab:disentangle_results}
\end{center}
\vspace{-2mm}
\end{table}

All models are trained in the same setting as the experiment shown in Figure \ref{fig:dsprites}, and are assessed by three disentangle metrics shown in Table \ref{tab:disentangle_results}. An improvement in the Z-Diff score and Factor score represents a lower variance of the inferred latent variable for fixed generative factors, whereas our increased SAP score corresponds with a tighter coupling between a single latent dimension and a generative factor. Compare to previous methods, our method is orthogonal (due to using a side objective) to most existing approaches. $\beta$-TCVAE \cite{chen2018isolating} improves $\beta$-VAE \cite{higgins2017beta} based on weighted mini-batches to stochastic training. Our Guided-$\beta$-TCVAE further improves the results in all three disentangle metrics.

\begin{table}
\begin{center}
\scalebox{0.7}{
\begin{tabular}{l | ccc}
\textbf{Model } & {$d_\z = 16$ $\downarrow$} &  {$d_\z = 32$ $\downarrow$}  &  {$d_\z = 64$ $\downarrow$} \\
\hline
\textbf{\textsc{VAE}   \cite{kingma2013auto}}
            & 2.92\%$\pm$0.12 & 3.05\%$\pm$0.42 & 2.98\%$\pm$0.14\\
\textbf{\textsc{$\beta$-VAE($\beta$=2)}\cite{higgins2017beta}} 
            & 4.69\%$\pm$0.18 & 5.26\%$\pm$0.22 & 5.40\%$\pm$0.33 \\
\textbf{\textsc{FactorVAE($\gamma$=5)} \cite{kim2018disentangling}} 
            & 6.07\%$\pm$0.05 & 6.18\%$\pm$0.20 & 6.35\%$\pm$0.48 \\
\textbf{\textsc{$\beta$-TCVAE ($\alpha$=1,$\beta$=5,$\gamma$=1)} \cite{chen2018isolating}} 
            & 1.62\%$\pm$0.07 & 1.24\%$\pm$0.05 & 1.32\%$\pm$0.09 \\
\hline
\textbf{\textsc{Guided-VAE (Ours)}}   
            & 1.85\%$\pm$0.08 & 1.60\%$\pm$0.08  & 1.49\%$\pm$0.06 \\            
\textbf{\textsc{Guided-$\beta$-TCVAE (Ours)}}   
            & \textbf{1.47\%$\pm$0.12 } & \textbf{1.10\%$\pm$0.03 } & \textbf{1.31\%$\pm$0.06} \\
\end{tabular}
}
\caption{\small \textbf{Representation Learning:} Classification error over unsupervised disentanglement methods on MNIST. [$\downarrow$ means lower is better]{\scriptsize \textsuperscript{$\dagger$} The 95 \% confidence intervals from 5 trials are reported.}}
\label{tab:classification-mnist-methods}
\end{center}
\vspace{-5mm}
\end{table}

\begin{figure*}[!htp]

\begin{center}
\scalebox{0.9}{
\begin{tabular}{ccc}

\includegraphics[width=0.31\textwidth]{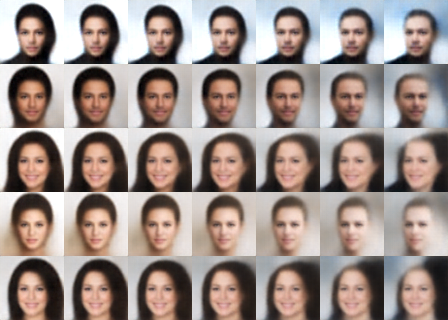}  &
\includegraphics[width=0.31\textwidth]{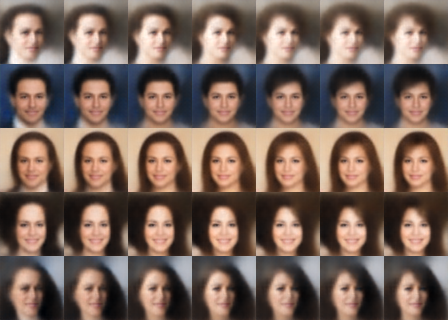} &
\includegraphics[width=0.31\textwidth]{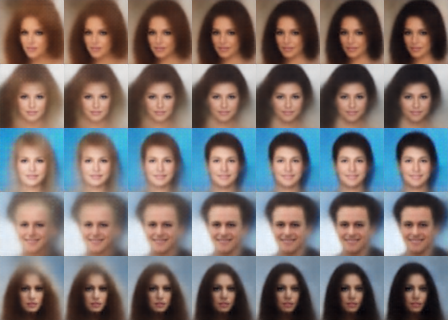}\\
(a) Bald  &
(b) Bangs &
(c) Black Hair\\
\vspace{+1mm}
\includegraphics[width=0.31\textwidth]{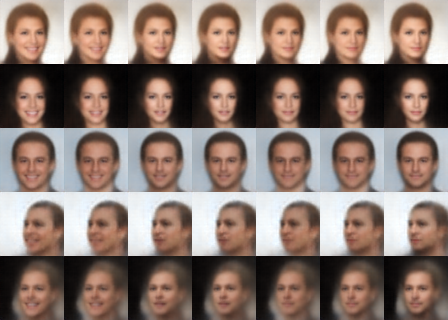} &
\includegraphics[width=0.31\textwidth]{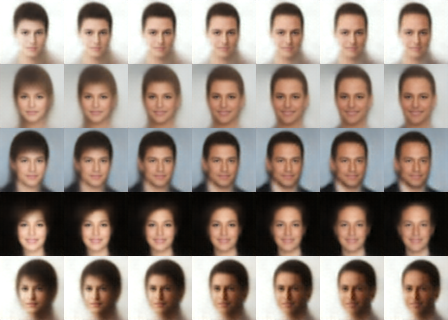}  &
\includegraphics[width=0.31\textwidth]{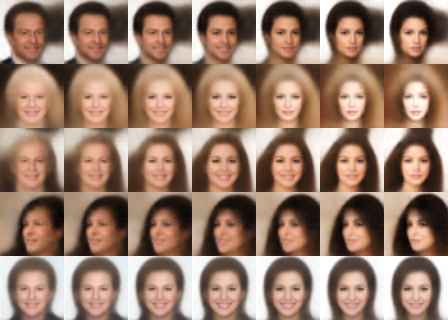}\\
(d) Mouth Slightly Open &
(e) Receding Hairlines &
(f) Young\\

\end{tabular}

}
\caption{\small\textbf{Latent factors learned by Guided-VAE on CelebA:} Each image shows the traversal results of Guided-VAE on a single latent variable which is controlled by the lightweight decoder using the corresponding labels as signal.}
\label{fig:celeba_appendix}
\end{center}
\end{figure*}

We further study representation transferability by performing classification tasks on the latent embedding of different generative models. Specifically, for each data point ($\mathbf{x}, y$), we use the pre-trained generative models to obtain the value of latent variable $\z$ given input image $\x$. Here $\z$ is a $d_{\z}$-dim vector. We then train a linear classifier $f(\cdot)$ on the embedding-label pairs $\{(\z, y)\}$ to predict the class of digits. For the Guided-VAE, we disentangle the latent variables $\z$ into deformation variables $\z_{def}$ and content variables $\z_{cont}$ with same dimensions (i.e. $d_{\z_{def}}=d_{\z_{cont}}$). We compare the classification errors of different models with multiple choices of dimensions of the latent variables in Table \ref{tab:classification-mnist-methods}. In general, VAE \cite{kingma2013auto}, $\beta$-VAE \cite{higgins2017beta}, and FactorVAE \cite{kim2018disentangling} do not benefit from the increase of the latent dimensions, and $\beta$-TCVAE \cite{chen2018isolating} shows  evidence that its discovered representation is more useful for classification task than existing methods. Our Guide-VAE achieves competitive results compare to $\beta$-TCVAE, and our Guided-$\beta$-TCVAE can further reduce the classification error to $1.1\%$ when $d_\z = 32$, which is $1.95\%$ lower than the baseline VAE.  

Moreover, we study the effectiveness of $\z_{def}$ and $\z_{cont}$ in Guided-VAE separately to reveal the different properties of the latent subspace. We follow the same classification task procedures described above but use different subsets of latent variables as input features for the classifier $f(\cdot)$. Specifically, we compare results based on the deformation variables $\z_{def}$, the content variables $\z_{cont}$, and the whole latent variables $\z$ as the input feature vector. To conduct a fair comparison, we still keep the same dimensions for the deformation variables $\z_{def}$ and the content variables $\z_{cont}$. Table \ref{tab:classification-mnist-disentanglement} shows that the classification errors on $\z_{cont}$ are significantly lower than the ones on $\z_{def}$, which indicates the success of disentanglement as the content variables should determine the class of digits. In contrast, the deformation variables should be invariant to the class. Besides, when the dimensions of latent variables $\z$ are higher, the classification errors on $\z_{def}$ increase while the ones on $\z_{cont}$ decrease, indicating a better disentanglement between deformation and content with increased latent dimensions.

\begin{table}
\begin{center}
\scalebox{0.65}{
\begin{tabular}{l | cccccc}
\textbf{Model } & $d_{\z_{def}}$  & {$d_{\z_{cont}}$}  & {$d_{\z}$} & {$\z_{def}\,\, Error$  $\uparrow$} &  {$\z_{cont}   \,\, Error$  $\downarrow$}  &  {$\z \,\,Error$ $\downarrow$}\\
\hline
\textbf{\textsc{Guided-VAE}} 
              &8 & 8 & 16   & 27.1\% & 3.69\% & 2.17\% \\
\,\,\,\,\,\,  &16 &16 & 32  & 42.07\% & 1.79\% & 1.51\% \\
\,\,\,\,\,\,  &32 & 32 & 64 & 62.94\% & 1.55\% & 1.42\% \\

\end{tabular}
}
\caption{\small \textbf{Classification on MNIST using different latent variables as features:} Classification error over Guided-VAE with different dimensions of latent variables [$\uparrow$ means higher is better, $\downarrow$ means lower is better]}
\label{tab:classification-mnist-disentanglement}
\end{center}
\vspace{-9mm}
\end{table}

\vspace{-2mm}
\subsection{Supervised Guided-VAE}

\subsubsection{Qualitative Evaluation}
We first present qualitative results on the CelebA dataset \cite{liu2015faceattributes} by traversing latent variables of attributes shown in Figure \ref{fig:Comparison_CelebA}  and Figure \ref{fig:celeba_appendix}. In Figure \ref{fig:Comparison_CelebA}, we compare the traversal results of Guided-VAE with $\beta$-VAE for two labeled attributes (gender, smile) in the CelebA dataset. The bottleneck size is set to 16 ($d_\z = 16$). We use the first two latent variables $z_1, z_2$ to represent the attribute information, and the rest $\z_{3:16}$ to represent the content information. During evaluation, we choose $z_t \in \{z_1, z_2\}$ while keeping the remaining latent variables $\z_t^{rst}$ fixed. Then we obtain a set of images through traversing $t$-th attribute (e.g., smiling to non-smiling) and compare them over $\beta$-VAE. In Figure \ref{fig:celeba_appendix}, we present traversing results on another six attributes.

$\beta$-VAE performs decently for the controlled attribute change, but the individual $\z$ in $\beta$-VAE is not fully entangled or disentangled with the attribute. We observe the traversed images contain several attribute changes at the same time. Different from our Guided-VAE, $\beta$-VAE cannot specify which latent variables to encode specific attribute information. Guided-VAE, however, is designed to allow defined latent variables to encode any specific attributes. Guided-VAE outperforms $\beta$-VAE by only traversing the intended factors (smile, gender) without changing other factors (hair color, baldness).
\vspace{-5mm}

\subsubsection{Quantitative Evaluation}

\begin{figure}[!htb]
\begin{center}
\vspace{-3mm}
\scalebox{0.9}{
\begin{tabular}{c}
\includegraphics[width=0.35\textwidth]{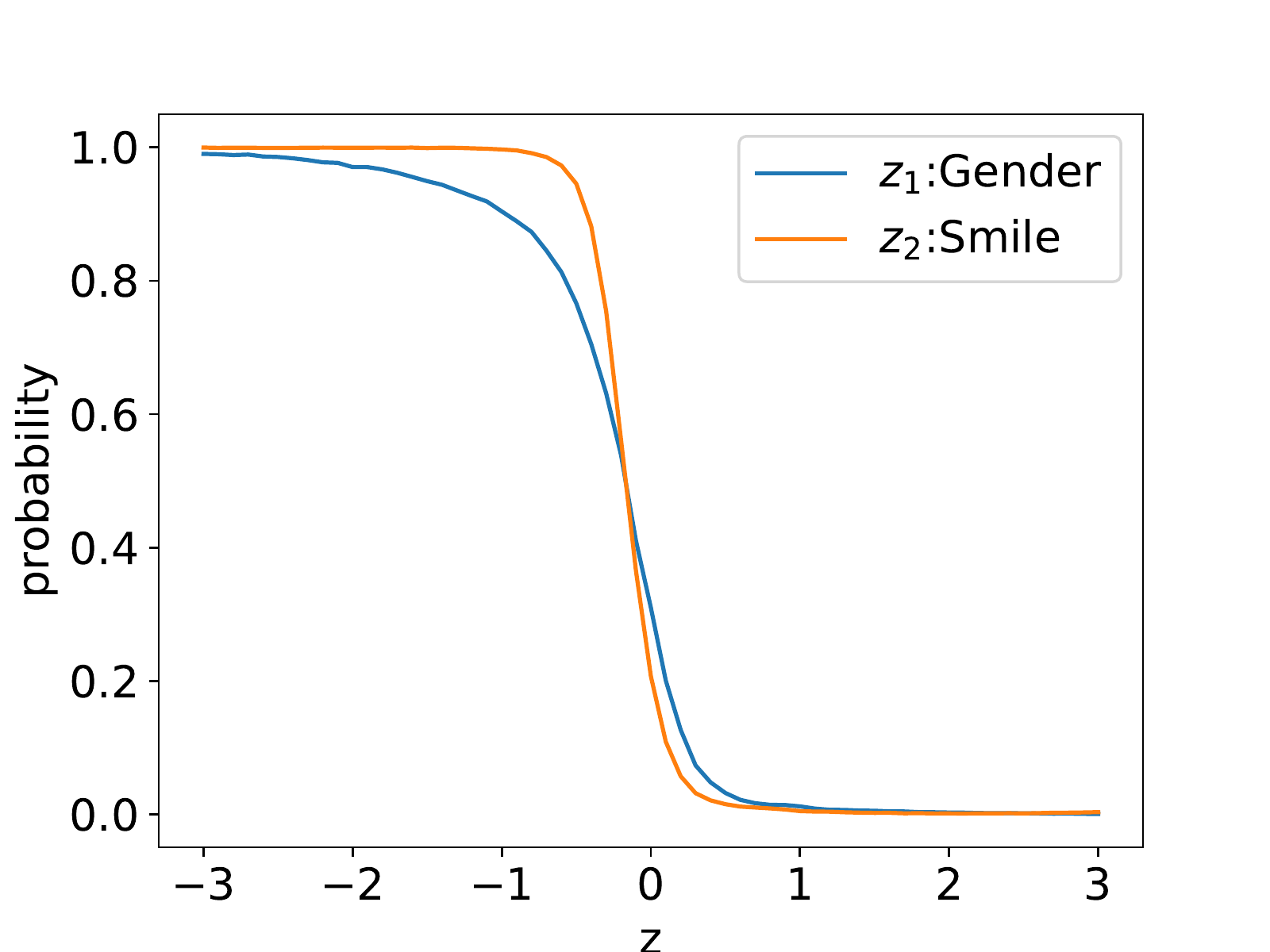}

\end{tabular}
}
\vspace{-1mm}
\caption{
\small Experts (high-performance external classifiers for attribute classification) prediction for being negatives on the generated images. We traverse $z_1$ (gender) and $z_2$ (smile) separately to generate images for the classification test.
}
\label{fig:supervised-quantitative}
\end{center}
\vspace{-7mm}
\end{figure}

We attempt to interpret whether the disentangled attribute variables can control the generated images from the supervised Guided-VAE. We pre-train an external binary classifier for $t$-th attribute on the CelebA training set and then use this classifier to test the generated images from Guided-VAE. Each test includes $10,000$ generated images randomly sampled on all latent variables except for the particular latent variable $z_t$ we decide to control. As Figure \ref{fig:supervised-quantitative} shows, we can draw the confidence-$z$ curves of the $t$-th attribute where $z=z_t\in [-3.0, 3.0]$  with $0.1$ as the stride length. For the gender and the smile attributes, it can be seen that the corresponding $z_t$ is able to enable ($z_t < -1$) and disable ($z_t > 1$) the attribute of the generated image, which shows the controlling ability of the $t$-th attribute by tuning the corresponding latent variable $z_t$. 

\vspace{-5mm}
\subsubsection{Image Interpolation}
\vspace{-2mm}

We further show the disentanglement properties of using supervised Guided-VAE on the CIFAR10 dataset. ALI-VAE borrows the architecture that is defined in \cite{ALI}, where we treat $G_z$ as the encoder and $G_x$ as the decoder. This enables us to optimize an additional reconstruction loss. Based on ALI-VAE, we implement Guided-ALI-VAE (Ours), which adds supervised guidance through excitation and inhibition shown in Figure \ref{fig:model}. ALI-VAE and AC-GAN \cite{acgan} serve as a baseline for this experiment. 

To analyze the disentanglement of the latent space, we train each of these models on a subset of the CIFAR10 dataset \cite{cifar10} (Automobile, Truck, Horses) where the class label corresponds to the attribute to be controlled. We use a bottleneck size of 10 for each of these models. We follow the training procedure mentioned in \cite{acgan} for training the AC-GAN model and the optimization parameters reported in \cite{ALI} for ALI-VAE and our model. For our Guided-ALI-VAE model, we add supervision through inhibition and excitation on $z_{1:3}$.  

\begin{figure}
\begin{center}
\begin{tabular}{c}
\vspace{-2mm}
\includegraphics[width=0.48\textwidth]{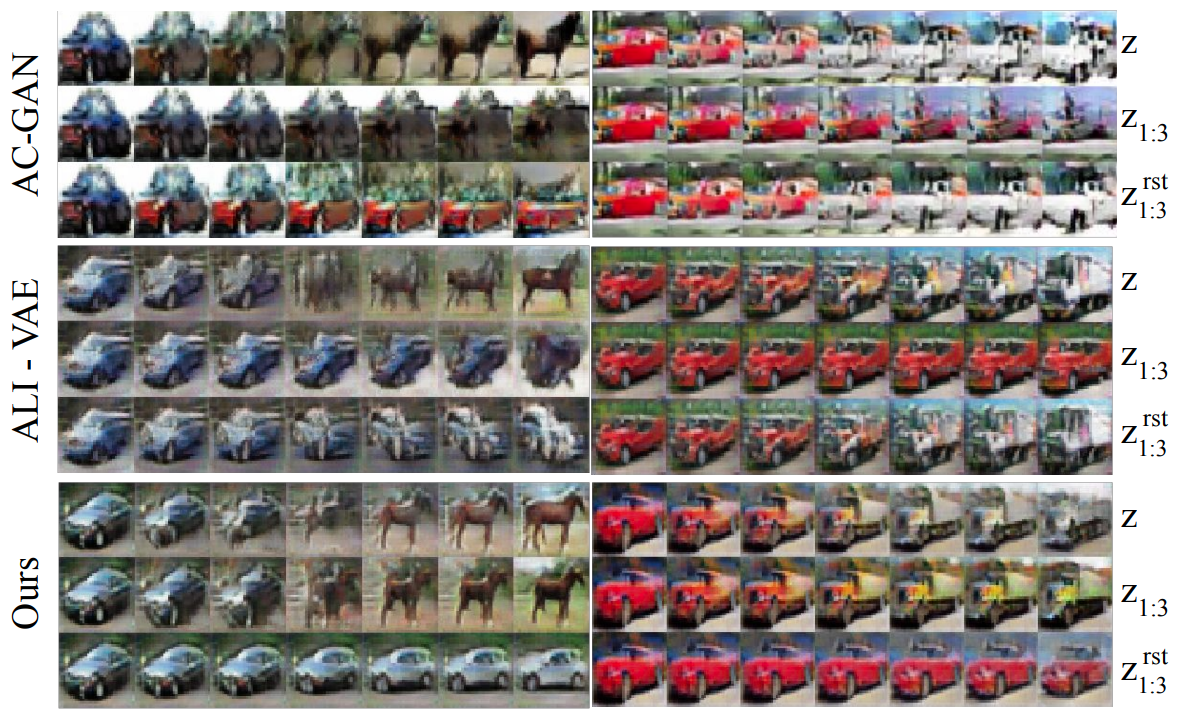}
\end{tabular}
\caption{\small Interpolation of images in $z$, $z_{1:3}$ and $z_{1:3}^{rst}$ for AC-GAN, ALI-VAE and Guided-ALI-VAE (Ours).
}
\label{fig:traverse-Interpolation}
\end{center}
\vspace{-8mm}
\end{figure}

\begin{table}
\begin{center}
\scalebox{0.75}{
\begin{tabular}{l | cc}
\textbf{Model }             & Automobile-Horse $\downarrow$   &  Truck-Automobile $\downarrow$   \\
\hline
\textbf{\textsc{AC-GAN} \cite{acgan}}    &   88.27             &   81.13 \\
\textbf{\textsc{ALI-VAE}} \textsuperscript{$\dagger$}   &   91.96             &   78.92 \\
\textbf{\textsc{Guided-ALI-VAE (Ours)}}      &   \textbf{85.43}             &  \textbf{72.31} \\
\end{tabular}
}
\caption{\small \textbf{Image Interpolation: } FID score measured for a subset of CIFAR10 \cite{cifar10} with two classes each. [$\downarrow$ means lower is better] \textsuperscript{$\dagger$} ALI-VAE is a modification of the architecture defined in \cite{ALI} }

\label{tab:cifar-fid}
\end{center}
\end{table}
To visualize the disentanglement in our model, we interpolate the corresponding $z$, $z_{t}$ and $z_{t}^{rst}$ of two images sampled from different classes. The interpolation here is computed as a uniformly spaced linear combination of the corresponding vectors. The results in Figure \ref{fig:traverse-Interpolation} qualitatively show that our model is successfully able to capture complementary features in $z_{1:3}$ and $z_{1:3}^{rst}$. Interpolation in $z_{1:3}$ corresponds to changing the object type. Whereas, the interpolation in $z_{1:3}^{rst}$ corresponds to complementary features such as color and pose of the object.

The right column in Figure \ref{fig:traverse-Interpolation} shows that our model can traverse in $z_{1:3}$ to change the object in the image from an automobile to a truck. Whereas a traversal in $z_{1:3}^{rst}$ changes other features such as background and the orientation of the automobile. We replicate the procedure on ALI-VAE and AC-GAN and show that these models are not able to consistently traverse in $z_{1:3}$ and $z_{1:3}^{rst}$ in a similar manner. Our model also produces interpolated images in higher quality as shown through the FID scores \cite{fid} in Table \ref{tab:cifar-fid}. 

\subsection{Few-Shot Learning}
Previously, we have shown that Guided-VAE can perform images synthesis and interpolation and form better representation for the classification task. Similarly, we can apply our supervised method to VAE-like models in the few-shot classification. Specifically, we apply our adversarial excitation and inhibition formulation to the Neural Statistician \cite{edwards2016towards} by adding a supervised guidance network after the statistic network. The supervised guidance signal is the label of each input. We also apply the Mixup method \cite{zhang2017mixup} in the supervised guidance network. However, we could not reproduce exact reported results in the Neural Statistician, which is also indicated in \cite{korshunova2018bruno}. For comparison, we mainly consider results from Matching Nets \cite{vinyals2016matching} and Bruno \cite{korshunova2018bruno} shown in Table \ref{tab:Omniglot}. Yet it cannot outperform Matching Nets, our proposed Guided Neural Statistician reaches comparable performance as Bruno (discriminative), where a discriminative objective is fine-tuned to maximize the likelihood of correct labels.

\begin{table}[!htb]
\begin{center}
\scalebox{0.8}{
\begin{tabular}{l | cc|cc}
\textbf{Model} & \multicolumn{2}{c|}{\textbf{5-way}} &  \multicolumn{2}{|c}{\textbf{20-way}} \\
\textbf{Omniglot } & \textbf{ 1-shot} & \textbf{5-shot} & \textbf{1-shot} & \textbf{ 5-shot}\\
\hline
\textbf{\textsc{Pixels} \cite{vinyals2016matching}} & 41.7\%  &63.2\% &26.7\% & 42.6\%\\
\textbf{\textsc{Baseline Classifier} \cite{vinyals2016matching}} & 80.0\%  &95.0\% &69.5\% & 89.1\%\\
\textbf{\textsc{Matching Nets} \cite{vinyals2016matching}} & 98.1\%  &98.9\% &93.8\% & 98.5\%\\
\textbf{\textsc{Bruno} \cite{korshunova2018bruno}} & 86.3\%  &95.6\% &69.2\% & 87.7\%\\
\textbf{\textsc{Bruno (discriminative)} \cite{korshunova2018bruno}} & 97.1\%  &99.4\% &91.3\% & 97.8\%\\

\hline
\textbf{\textsc{Baseline} } & 97.7\%  &99.4\% &91.4\% & 96.4\%\\
\textbf{\textsc{Ours (discriminative)}} &97.8\% &99.4\%  &92.1\% &96.6\%\\
\end{tabular}
}
\caption{\small \textbf{Few-shot classification:} Classification accuracy for a few-shot learning task on the Omniglot dataset.}
\label{tab:Omniglot}
\vspace{-5mm}
\end{center}
\end{table}

\vspace{-4mm}
\section{Ablation Study}
\label{others}

\subsection{Deformable PCA}
\begin{figure}[!htb]
\begin{center}
\scalebox{0.7}{
\begin{tabular}{c}
\includegraphics[width=0.52\textwidth]{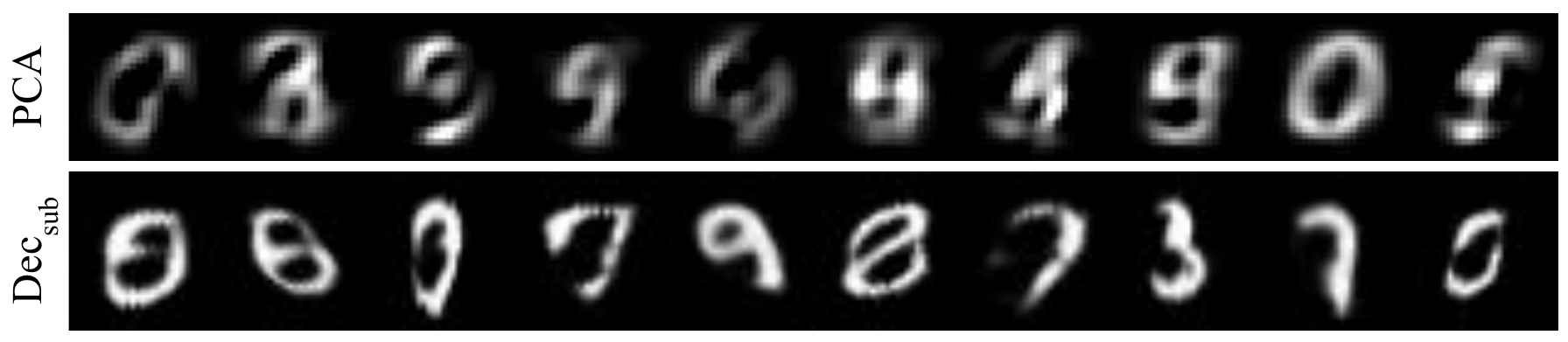}\\
 
\end{tabular}
\vspace{-2mm}
}

\caption{\small (Top) Sampling Result Obtained from PCA (Bottom) Sampling Result obtained from learned deformable PCA (Ours)}
\label{fig:pca}
\end{center}
\end{figure}

In Figure \ref{fig:pca}, we visualize the sampling results from PCA and $Dec_{sub}$. By applying a deformation layer into the PCA-like layer, we show deformable PCA has a more crispy sampling result than vanilla PCA. 

\vspace{-6mm}
\subsection{Guided Autoencoder}
\vspace{-2mm}

To further validate our concept of ``guidance'', we introduce our lightweight decoder to the standard autoencoder (AE) framework. We conduct MNIST classification tasks using the same setting in Figure \ref{tab:classification-mnist-methods}. As Table \ref{tab:classification-mnist-ae} shows, our lightweight decoder improves the representation learned in autoencoder framework. Yet a VAE-like structure is indeed not needed if the purpose is just reconstruction and representation learning. However, VAE is of great importance in building generative models. The modeling of the latent space of ${\bf z}$ with e.g., Gaussian distributions is again important if a probabilistic model is needed to perform novel data synthesis (e.g., the images shown in Figure \ref{fig:Comparison_CelebA}  and Figure \ref{fig:celeba_appendix}).

\begin{table}[h!]
\begin{center}
\scalebox{0.8}{
\begin{tabular}{l | ccc}
\textbf{Model } & {$d_\z = 16$ $\downarrow$} &  {$d_\z = 32$ $\downarrow$}  &  {$d_\z = 64$ $\downarrow$} \\
\hline
\textbf{\textsc{Auto-Encoder (AE)}}
              & \textbf{1.37}\%$\pm$0.05 & 1.06\%$\pm$0.04 & 1.34\%$\pm$0.04 \\
\textbf{\textsc{Guided-AE (Ours)}}
              & 1.46\%$\pm$0.06 & \textbf{1.00}\%$\pm$0.06 & \textbf{1.10}\%$\pm$0.08 \\
\end{tabular}
}
\caption{\footnotesize Classification error over AE and Guided-AE on MNIST.}
\label{tab:classification-mnist-ae}
\end{center}
\vspace{-7mm}
\end{table}

\subsection{Geometric Transformations}
\vspace{-2mm}

\begin{figure}[!htb]
\begin{center}
\scalebox{0.9}{
\begin{tabular}{cc}
\includegraphics[width=0.23\textwidth]{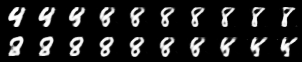}&
\includegraphics[width=0.23\textwidth]{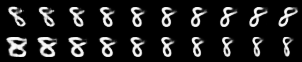}
\\ 
\includegraphics[width=0.23\textwidth]{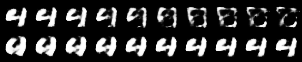}&
\includegraphics[width=0.23\textwidth]{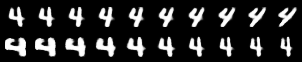}
\\
\small{(a) Unsupervised Guided-VAE} & \small{(b) Unsupervised Guided-VAE}\\
\small{without Geometric Guidance} & \small{with Geometric Guidance}\\

\includegraphics[width=0.23\textwidth]{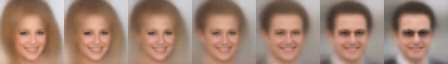}
& \hspace{-2mm}
\includegraphics[width=0.23\textwidth]{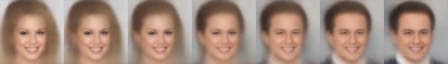}
\\ 
\includegraphics[width=0.23\textwidth]{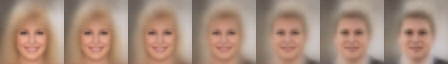}
& \hspace{-2mm}
\includegraphics[width=0.23\textwidth]{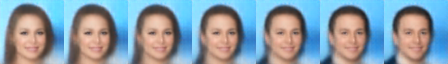}
\\ 
\small{(c) Supervised Guided-VAE} & \small{(d) Supervised Guided-VAE}\\
\small{without Inhibition} & \small{with Inhibition}\\
\end{tabular}
}
\vspace{1mm}
\caption{\small{Ablation study on Unsupervised Guided-VAE and Supervised Guided-VAE}}
\label{fig:ablation}
\end{center}
\vspace{-5mm}
\end{figure}

We conduct an experiment by excluding the geometry-guided part from the unsupervised Guided-VAE. In this way, the lightweight decoder is just a PCA-like decoder but not a deformable PCA. The setting of this experiment is exactly the same as described in  Figure \ref{fig:mnist}. The bottleneck size of our model is set to 10 of which the first two latent variables $z_1, z_2$ represent the rotation and scaling information separately. As a comparison, we drop off the geometric guidance so that all 10 latent variables are controlled by the PCA-like light decoder. As shown in Figure \ref{fig:ablation} (a) (b), it can be easily seen that geometry information is hardly encoded into the first two latent variables without a geometry-guided part.

\vspace{-2mm}
\subsection{Adversarial Excitation and Inhibition}
\vspace{-2mm}
We study the effectiveness of adversarial inhibition using the exact same setting described in the supervised Guided-VAE part. As shown in Figure \ref{fig:ablation} (c) and (d), Guided-VAE without inhibition changes the smiling and sunglasses while traversing the latent variable controlling the gender information.
This problem is alleviated by introducing the excitation-inhibition mechanism into Guided-VAE.

\vspace{-2mm}
\section{Conclusion}
\vspace{-2mm}
In this paper, we have presented a new representation learning method, guided variational autoencoder (Guided-VAE), for disentanglement learning. Both unsupervised and supervised versions of Guided-VAE utilize lightweight guidance to the latent variables to achieve better controllability and transparency. Improvements in disentanglement, image traversal, and meta-learning over the competing methods are observed. Guided-VAE maintains the backbone of VAE and it can be applied to other generative modeling applications. \\
\hspace{-1mm}{\bf Acknowledgment}. \small{This work is funded by NSF IIS-1618477, NSF IIS-1717431, and Qualcomm Inc. ZD is supported by the Tsinghua Academic Fund for Undergraduate Overseas Studies. We thank Kwonjoon Lee, Justin Lazarow, and Jilei Hou for valuable feedbacks.}

{\small
\bibliographystyle{ieee_fullname}
\bibliography{egbib}
}

\end{document}

%% file: math_commands.tex
\usepackage{amsmath,amsfonts,bm}

\def\1{\bm{1}}

\DeclareMathAlphabet{\mathsfit}{\encodingdefault}{\sfdefault}{m}{sl}
\SetMathAlphabet{\mathsfit}{bold}{\encodingdefault}{\sfdefault}{bx}{n}

\def\gL{{\mathcal{L}}}

\newcommand{\KL}{D_{\mathrm{KL}}}

